# iDeLog: Iterative Trajectory and Velocity Sigma-Lognormal Parameter Extraction

Miguel A. Ferrer, Moises Diaz, Cristina Carmona-Duarte, Réjean Plamondon

*Abstract*— Kinematic theory of rapid movements and its associated Sigma-lognormal model has been extensively used in a wide variety of application. While the physical and biological meaning of the model has been an unquestionable progress, there are some open issues in the current algorithms that calculate the Sigma-Lognormal parameters of complex and long movements. This paper proposes a novel algorithm to work out the parameters of the Sigma-Lognormal model, called iDeLog. Its main novelty relies on adjusting the trajectory and velocity jointly obtaining a better reconstruction of an original human movement. Specifically, once the initial set of lognormals has been automatically worked out, iDeLog iteratively moves the virtual targets points to improve the adjustment between original and reconstructed trajectory. New characteristics as the possibility of using Clothoids curves instead of circumference arcs to link virtual target points or using different functions to adjust the velocity bell of each stroke are also available. In the experimental part of this paper, iDeLog is compared with ScriptStudio, obtaining promising performances. Moreover, iDeLog can be freely downloaded for research proposes.

## I. INTRODUCTION

Human movement modeling is of great interest for designing intelligent systems relying on the understanding of the fine motor control.

There exist many theories that have tried to approach the velocity profile of human movement in general and handwriting in particular. Specifically, in [1] are mentioned models relying on neural networks [2][3][4][5], equilibrium point models [6][7][8], behavioral models [9][10][11], coupled oscillator models [12][13][14], differential equation models [15], kinematic models [16][17][18], and models exploiting minimization principles [19][20] such as minimization of the acceleration [21][22], or the energy [23], or the time [24][25][26], or the jerk [27][28], or the snap [29], or the torque changes [30] or the sensory-motor noise [31]. Finally, many models exploit the properties of various mathematical functions to reproduce human movements: exponentials [32], second order systems [33][34], Gaussians [35], beta functions [36], splines [37] and trigonometrical functions [38].

Among the models which provide analytical representations, the kinematic theory of rapid human movements [16][17][39][40][18] and its Delta and Sigma–lognormal models have been extensively used to explain most of the basic phenomena reported in classical studies on human motor control [41] and to study several factors involved in the fine motricity [42][43][44][45][46][47][48][49].

Apart from these fundamental studies, the theory has been used, directly or indirectly, in many practical applications like the development of Automatic Signature Verification (ASV) systems [49][50], the generation of duplicated signatures to improve the training [51][52] and the improvement of the forgery detection [53]. Moreover, it has been used to generate synthetic Western [54][55][56][57] and Indian [58][59] signatures. Beyond the ASV field, this model has been successfully applied to study and understand the evolution acquiring the handwriting skills [60], the development of tools to help children learning handwriting [61], for neuromuscular health care [62], as well as of biomedical set ups to detect fine motor control problems associated with brain strokes [63][64], Parkinson disease [65][66] and to characterize the turn cranio-caudal signature [67]. Lately, the Kinematic theory for rapid movements has also been applied at many different areas such as to generate handwriting [68], CAPTCHA [69][70] [71], graffiti design [72][73], analysis of mouse movements [74], gestures generation [75][76][77][78] and to study articulation in voice processing [79][80], among other.

As can be seen, the Sigma lognormal model have represented a breakthrough due to its feasibility and reliability to describe a wide variety of human movements. The main advantage of this model is, probably, the consideration of physical body features such as the neuromuscular system responsible for the production of human movements, which reflects some personal characteristics difficult to impersonate.

To work out the Sigma-Lognormal parameters, the Robust Xzero algorithm implemented by the ScriptStudio application was proposed in 2007 [46][47]. Since its first version, it was quickly widespread, and several improvements have been published, for instance in [81].

With hindsight, after 10 years of intensive use in several fields, there are room for some improvements in the algorithmic implementation of Sigma-Lognormal model. For instance, it is well known the trajectory drift when reconstructing long movements with Script Studio. This issue has been addressed by changing the stop criteria and segmenting the long

This study was funded by the Spanish government's MIMECO TEC2016-77791-C4-1-R research project and European Union FEDER program/funds.

Miguel A. Ferrer, Moises Diaz and Cristina Carmona-Duarte are with the Instituto Universitario para el Desarrollo Tecnológico y la Innovación en Comunicaciones, Universidad de Las Palmas de Gran Canaria, Campus de Tafira, Las Palmas de Gran Canaria, Spain. Emails: mferrer@idetic.eu, mdiaz@idetic.eu, ccarmona@idetic.eu

Moises Diaz is also with Mid-Atlantic University, Las Palmas de Gran Canaria, Spain. Email: mdiazc@unidam.es

Réjean Plamondon is with École Polytechnique, Université de Montréal, Montréal, P.Q., Canada. Email: rejean.plamondon@polymtl.ca

handwriting in smaller pieces and reconstruct each of them independently [81].

In this paper we reformulate from the scratch the algorithm to work out the Sigma-Lognormal parameters. The novelty of the new algorithm is a joint optimization of the reconstructed trajectory and velocity. In this way, iDeLog firstly decides the number of lognormal from the number of minima in the velocity profile. Next, the velocity profile is decomposed as a weighted sum of lognornals. The virtual Target Points and angles of each circumference are calculated from the corresponding position of the minima in the 8-connected trajectory. In this way an initial reconstruction is obtained which is optimized iteratively moving the position of the virtual target points with the ensuing changes of the angles and lognormal parameters. It is worthy pointing out that not new lognormals are added to improve the reconstruction rate. A block diagram of the algorithm in the case of signatures is shown in Figure 1.

The remainder of the paper is as follows: Section II reviews briefly the Sigma-Lognormal model and the Sigma-Lognormal extractor behind ScriptStudio. Section III is devoted to develop iDeLog algorithm whilst Section IV introduces additional issues of iDeLog algorithm such as linking virtual points by Clothoid curves or using different bell functions to approach the velocity bells. The evaluation and related discussions are described in Section V while Section VI concludes the article.

## II. SIGMA-LOGNORMAL MODEL: A BRIEF REVIEW

### A. The Sigma-Lognormal Model

The Kinematic theory of rapid movements [16], from which the Sigma-Lognormal model was developed [47], consider complex movements as overlapping strokes. Each stroke has a lognormal shaped velocity profile $v_j(t)$ defined as:

$$v_j\left(t; t_{0_j}, \mu_j, \sigma_j^2\right) = D_j \Lambda\left(t; t_{0_j}, \mu_j, \sigma_j\right) = \frac{D_j}{\sigma_j \sqrt{2\pi}(t - t_{0_j})} exp\left\{-\frac{\left[ln\left(t - t_{0_j}\right) - \mu_j\right]^2}{2\sigma_j^2}\right\} \quad (1)$$

where $t$ is the basis of time, $t_{0_j}$ the time of stroke occurrence, $D_j$ the amplitude of the stroke, $\mu_j$ the stroke time delay on a logarithmic time scale and $\sigma_j$ the stroke response time.

The overlapping of these lognormals can produce acomplex trajectory, known as trajectory plan. A trajectory plan consists of a sequence of virtual target points linked together by circular arcs. Each arc movement is produced as response of the motor system to a set of rhythmic commands from the cerebellum. The overlap in time of this movements results as:

$$\vec{v}(t) = \begin{bmatrix} v_x(t) \\ v_y(t) \end{bmatrix} = \sum_{j=1}^{N} D_j \begin{bmatrix} cos\phi_j(t) \\ sin\phi_j(t) \end{bmatrix} D_j \Lambda\left(t; t_{0_j}, \mu_j, \sigma_j\right) \quad (2)$$

where $N$ is the number of lognormal strokes and $\phi_j(t)$ is the angular position.

$$\phi_j(t) = \theta_{s_j} + \frac{\theta_{e_j} - \theta_{s_j}}{2}\left[1 + erf\left(\frac{ln\left(t - t_{0_j}\right) - \mu_j}{\sigma_j \sqrt{2}}\right)\right] \quad (3)$$

being $\theta_{e_j}$ and $\theta_{s_j}$ the starting angle and the end angle of the arc that link the two virtual target points corresponding to the $j^{th}$ stroke. Note that this formula describes the swept from $\theta_{s_j}$ to $\theta_{e_j}$ in a lognormal timing. Finally, the trajectory is worked out as:

$$\vec{s}(t) = \begin{bmatrix} x_r(t) \\ y_r(t) \end{bmatrix} = \sum_{j=1}^{N} \frac{D_j}{\theta_{e_j} - \theta_{s_j}} \begin{bmatrix} sin\phi_j(t) - sin\theta_{s_j} \\ -cos\phi_j(t) + cos\theta_{s_j} \end{bmatrix} \quad (4)$$

This formula converts angles into circular arcs and overlap them. Specifically, the $j^{th}$ term of the summation represents the arc of circumference that link the virtual target points $tp_{j-1}$ and $tp_j$. The radius of such circumference is $D_j/\left(\theta_{e_j} - \theta_{s_j}\right)$ and $D_j$ coincides with the length of the arc. In this case, the virtual target points are defined by:

$$tp_j = tp_{j-1} + \frac{D_j}{\theta_{e_j} - \theta_{s_j}} \begin{bmatrix} sin\phi_j(T) - sin\theta_{s_j} \\ -cos\phi_j(T) + cos\theta_{s_j} \end{bmatrix} \quad (5)$$

being $T$ the temporal length of the signature.

### B. Strokes Extraction

It has been shown the robustness and flexibility of ScriptStudio to automatically extract the Sigma-lognormal parameters of rapid and complex human movements.

As it is a velocity-based algorithm, the procedure starts working out the magnitude of the velocity profile. To enhance the quality of the original velocity signal, it is preprocessed. Briefly, it is re-sampled to 200Hz through cubic splines and smoothed by a Chebyshev filter [47][81].

The strokes are extracted following the next iterative algorithm:

1. The strokes are identified in the velocity magnitude profile according to the lobes. In short, a lognormal is defined by its maximum $p_3$, and their inflection points $p_2$ and $p_4$ which are found by looking for changes in the sign of the curvature before and after the maximum, respectively.
2. The values of $t_{0_j}$, $\mu_j$ and $\sigma_j$ are obtained from $p_2$, $p_3$ and $p_4$. If the values of $\mu_j$ and $\sigma_j$ are out of the expected range, the values $p_2$ and $p_4$ are recalculated. As noise can generate false positive strokes, if the area and the maximum of the lognormal is lower than a threshold, the lognormal is discarded.
3. Once identified a stroke, its velocity is removed from the original velocity profile. This subtraction allows to uncover potential strokes whose local maxima was hidden by a faster neighbor lognormal.
4. The similarity between the preprocessed original velocity $v_o(t)$ and reconstructed velocity profile $v_r(t)$ is work out by means of the Signal-to-noise-ratio defined as:

$$SNR_v = 10log\left(\frac{\int_{t=0}^{T} v_o(t)^2 \, dt}{\int_{t=0}^{T}(v_o(t) - v_r(t))^2 dt}\right) \quad (6)$$

$T$ being its temporal length.

5. Go to point 1 considering the subtracted velocity as input velocity unless the $SNR_v$ was bigger than a threshold or the current number of strokes was bigger than a maximum or there were not any set of $p_2$, $p_3$ and $p_4$ points left.

### C. Stroke Parameters Estimation

The Robust XZERO algorithm [45] was used to estimate the lognormal parameters that approach the velocity profile of each stroke. Let $s_j$ be a stroke and $p_2$, $p_3$ and $p_4$ its characteristic points, which occurs at times $\{t_\alpha\}_{\alpha=2}^4$. Then their corresponding velocity values are $\{v_o(t_\alpha)\}_{\alpha=2}^4$. The lognormal parameters $\tau_j$, $\mu_j$, $\sigma_j$ and $D_j$ that fit $\{v_o(t_\alpha)\}_{\alpha=2}^4$ can be derived using different two-element combinations of the lognormal characteristic points as:

$$\sigma^2 = \begin{cases} -2 - 2\log r_{\alpha\beta} - \dfrac{1}{2\log r_{\alpha\beta}} & if\ \alpha = 2, \beta = 3 \\ -2 + 2\sqrt{1 + \log^2 r_{\alpha\beta}} & if\ \alpha = 2, \beta = 4 \\ -2 - 2\log r_{\beta\alpha} - \dfrac{1}{2\log r_{\beta\alpha}} & if\ \alpha = 3, \beta = 4 \end{cases} \quad (7)$$

where log is the natural logarithm and $r_{\alpha\beta} = \|\overrightarrow{v_o}(t_\alpha)\|/\|\overrightarrow{v_o}(t_\beta)\|$.

$$\mu = \log\left(\frac{t_\alpha - t_\beta}{e^{-a_\alpha} - e^{-a_\beta}}\right) \quad (8)$$

$$t_0 = t_\alpha - e^{\mu - a_\alpha} \quad (9)$$

$$D = \|\overrightarrow{v_o}(t_\alpha)\|\sigma\sqrt{2\pi}\exp\left(\mu + \frac{a_\alpha^2}{2\sigma^2}\right) - a_\alpha \quad (10)$$

where $\alpha, \beta \in \{2,3,4\}, \alpha < \beta$ and

$$a_i = \begin{cases} \dfrac{3}{2}\sigma^2 + \sigma\sqrt{\dfrac{\sigma^2}{4} + 1} & if\ i = 2 \\ \sigma^2 & if\ i = 3 \\ \dfrac{3}{2}\sigma^2 + \sigma\sqrt{\dfrac{\sigma^2}{4} + 1} & if\ i = 4 \end{cases} \quad (11)$$

The parameters are computed using all possible combinations. The set of parameters which minimizes the least-square error is kept as solution.

### D. Angle Estimation

The XZERO algorithm works out the start $\theta_s$ and end $\theta_e$ angles that link two virtual target points as follows:

$$\theta_s = \phi(t_3) - \Delta\phi(d(t_3) - d(t_1)) \quad (12)$$
$$\theta_e = \phi(t_3) - \Delta\phi(d(t_5) - d(t_3)) \quad (13)$$

where

$$\Delta\phi = \frac{\phi(t_4) - \phi(t_2)}{d(t_4) - d(t_2)} \quad (14)$$

$$d(t_i) = \begin{cases} 0 & if\ i = 1 \\ \dfrac{D}{2}\left[1 + erf\left(-a_i/\sigma\sqrt{2}\right)\right] & if\ i = 2,3,4 \\ D & if\ i = 5 \end{cases} \quad (15)$$

### III. IDELOG: NOVEL SIGMA-LOGNORMAL EXTRACTOR

As the ScriptStudio procedure based on Robust XZERO algorithm to work out the parameters of the Sigma-Lognormal model is velocity based, instead of trajectory-based, it is prone to spatial deviation as the original movement run longer. It is due to, according to Eq. 2, the velocity of the original movement is the overlapped sum of the velocity of each stroke. As a result, errors in the velocity estimation are propagated over the entire movement, resulting on an increased spatial deviation [82].

To improve the shape fitting, a dual velocity/trajectory optimization would be more appropriate. Some efforts have been carried out in this direction. For instance, in [81] the original signal is chopped into smaller pieces. They consider each piece as generated from a different action plan, therefore the extraction can be performed independently. Specifically, the original signal was based on dynamic handwriting, which was divided into pen-strokes, preserving only those that are pen-downs. Obviously, the shorter the length of the signal to be fitted, the shorter error propagation along the signal and, therefore, better improvement in the shape fitting.

In this section we propose a thoroughly new algorithm get off the ground, that overcomes the error propagation by fitting the shape and velocity simultaneously. This algorithm is called by us iDeLog which stands for iterative Decomposition in Lognormals.

### A. Stroke Extraction

As in the case of the ScriptStudio, the strokes are estimated from the magnitude of the velocity. In the case of iDeLog algorithm, it includes an option for preprocessing the original signal or not. This option allows to smooth the input signal in the same way that ScripStudio: cubic spline interpolation. As a difference in the preprocessing, the sample frequency is not changed in iDeLog.

Once carried out such preprocessing, if any, the strokes are identified as the velocity bells between velocity minima. Specifically, let the time of the velocity minima be $\{t_{min_j}\}_{j=0}^N$, where $N$ is the number of strokes, $t_{min_0} = 0$, and $t_{min_N} = T$ the temporal length of the original rapid movement. In this case, the velocity bell which correspond to stroke $j$ is defined as:

$$v_{o_j}(t) = \begin{cases} v_o(t) & t_{min_{j-1}} \leq t \leq t_{min_j} \\ 0 & otherwise \end{cases}, 1 < j < N \quad (16)$$

In the trajectory, the velocity minima corresponds to the salient points $sp_j, j = 0, \dots, N$, where $sp_0$ is the first sample, $sp_N$ is the last sample and $sp_j$ is the sample $t_{min_j} * f_m$ being $f_m$ the sampling frequency.

The next section works out the lognormal parameters that better approach the velocity bell of each stroke.

### B. Stroke parameters Estimation

This section approach each velocity bell by a lognormal function. ScriptStudio performs this step by working out the lognormal paramaters parameters $t_{0_j}$, $\mu_j$, $\sigma_j$ and $D_j$ that fits the velocity bell through XZERO algorithm as described above.

The approach in iDeLog is quite different as the estimation of the lognormal parameters take into account the stroke trajectory.

#### 1) Estimation of $t_{0_j}$

It is well-known that the movement action is issued in the cortex, passes through the Basal Ganglia that decodes the message in order to activate the different pool of muscles to carry out such movement. In a well-learned movement, the time between the movement action is issued and the movement is performed should be rather similar for each stroke. In our estimation procedure, this tramission time is approched by $t_{min,j-1} - t_{0_j}$ which is around 0.5 seconds [83].

*2) Initial Estimation of $\mu_j$ and $\sigma_j^2$*

IDeLog tries to fit the velocity bell $v_{o_j}(t)$ by the lognormal $v_j\left(t; t_{0_j}, \mu_j, \sigma_j^2\right)$ by means of nonlinear least-squares minimization. The function to minimize is:

$$\hat{\mu}_j, \hat{\sigma}_j^2 = \underset{\mu_j, \sigma_j^2}{\operatorname{argmin}} \int_{t=0}^{T} \left| v_j\left(t; t_{0_j}, \mu_j, \sigma_j^2\right) - v_{n_j}(t) \right|^2 dt \quad (17)$$

where:

$$v_{n_j}(t) = v_{o_j}(t) / \int_{t=0}^{T} v_{o_j}(t) dt \quad (18)$$

as the area of the lognormal function is equal to 1.

The minimization is performed by means of a Levenberg-Marquardt algorithm (LMA), which is used in many software applications for solving generic curve-fitting problems. However, the LMA finds only a local minimum, which is not necessarily the global minimum. The minimum found depends on the initial values of $\mu_j$ and $\sigma_j$ which are heuristically set up to $\mu_j = -0.5$ and $\sigma_j = 0.05$

Obviously, there is not an only solution for the lognormal $v_j\left(t; t_{0_j}, \mu_j, \sigma_j^2\right)$. Figure 2 shows different solution for the same velocity bell in function of $t_{min,j-1} - t_{0_j}$. All of them are valid solution from the computational point of view. From the biological point of view, it is expected fairly constant values of $\mu_j$ and $\sigma_j$ along the whole movement as the used motor system has not presumibly changed.

*C. Location of the Initial Virtual Target Points $tp_j$*

The virtual targets points are related with the velocity minima. A virtual target point mark the end of a stroke and the start of the next one: the overlap of the speed down to end a stroke (tail of the lognormal) with the speed up to start the next stroke (initial rise of the lognormal) turns in a velocity minimum.

From the spatial point of view, the trajectory is build up overlapping the arc of circumference that link consecutive virtual target points. So, a virtual target point involves two arc of circumferences. The one that comes from the previous virtual target point that arrives with an end angle and the one that departs toward the next virtual target point with a start angle. The overlap between the circumference arcs generate a trajectory with a higher curvature or salient point around the virtual target point.

As the virtual target points refers to spatial trajectory, iDeLog estimates the initial virtual target points directly from the original trajectory. Each salient point $sp_j$ is associated to a virtual target points $tp_j, j = 0, ..., N$. In iDeLog, $tp_0 = sp_0$ and $tp_N = sp_N$ which are the first and last point of the trajectory. For $j = 1, ..., N-1$, the virtual target points $tp_j$ is calculated using $sp_{j-1}$, $sp_j$ and $sp_{j+1}$, which form a triangle. The initial virtual target point is located on the median of the vertex $sp_j$, which is a straight line through the vertex $sp_j$ and the midpoint $(sp_{j-1} + sp_{j+1})/2$ of the opposite side, at a distance $dtp_j$ from the vertex $sp_j$ defined as:

$$dtp_j = Dtp_j cos(\varphi_j/2) \quad (19)$$

where $Dtp_j$ is the distance between the vertex $sp_j$ and the midpoint of the opposite side and $\varphi_j$ is the angle of the vertex $sp_j$. In this way, The closer is the angle of the vertex $sp_j$, the further is $tp_j$ from $sp_j$. An example of this procedure is shown in Figure 3-a.

*D. Estimation of Initial $\theta_{s_j}$ and $\theta_{e_j}$*

The angles of the circumferences that link virtual target points are defined with their start $\theta_{s_j}$ and end $\theta_{e_j}$ angles for $j = 1, ..., N$ being $N$ the number of strokes. As it is a spatial characteristic of the signature, iDeLog estimates these parameters from the original spatial trajectory.
They are calculated as follows:
1. The middle point $mp_j$ of the trajectory of the $j^{th}$ stroke between the salient points $sp_{j-1}$ and $sp_j$ is worked out. As middle point we meant the same distance on the trajectory from $mp_j$ to $sp_{j-1}$ and $sp_j$.
2. A circumference that passes by these three points is obtained.
3. The angle $\theta_{s_j}$ is computed as the angle of the tangent of the circumference in $sp_{j-1}$. Then, the angle $\theta_{e_j}$ is obtained as the angle of the tangent of the circumference in $sp_j$.

An illustration of this procedure is shown in Figure 3-b.

*E. Reconstructed Spatial Trajectory*

As a result of the above procedure, we have an initial gross estimation of the parameters of the Sigma-Lognormal model $t_{0_j}, \mu_j, \sigma_j^2, tp_j, \theta_{s_j}$ and $\theta_{e_j}, j = 1, ..., N$ for the original samples $\{x_o(t), y_o(t)\}, 0 < t < T$. From these parameters, we calculate the parameter $D_j$ and work out the reconstructed spatial trajectory $T_r(t) = \{x_r(t), y_r(t)\}$ following equations (3) and (4). The salient points of the reconstructed spatial trajectory $spr_j, \forall j \in 0, ..., N$ are obtained through the minima of the reconstructed velocity which is the derivate of the reconstructed trajectory.

*1) Estimation of $D_j$*

The value of the lognormal amplitude $D_j$ describes the amplitude of the movement and it is defined without ambiguity by the position of the virtual target points $tp_{j-1}, tp_j, \theta_{s,j}$ and $\theta_{e,j}$. It is calculated as:

$$D_j = r_j\left(\theta_{e_j} - \theta_{s_j}\right), \quad j \in 1, ... N \quad (19)$$

where $r_j$ is the radious of the circumference that goes from $tp_{j-1}$ to $tp_j$. To work out $r_j$, we first calculate the center of the circumference as the intersection of the line that transverse $tp_{j-1}$ with slope $-1/\tan \theta_{s_j}$ and the line that transverse $tp_j$ with slope $-1/\tan \theta_{e_j}$. Then, $r_j$ is the distance from the center of the circumference to either $tp_{j-1}$ or $tp_j$.

Note than the Sigma-Lognormal model overdefines the circuference with five parameters: $D_j, tp_{j-1}, tp_j, \theta_{s,j}$ and $\theta_{e,j}$, when a circumference only requires three parameters. It will allow some of the extensions of the Sigma-lognormal model, which will be defined later.

An example of a recovered spatial trajectory with these initial gross parameters is shown in Figure 4.

*F. Refinement algorithm*

The basic idea of the refinement is to optimize the position of the virtual target points $tp_j, j = 1, ..., N-1$ to improve both the reconstructed trajectory and velocity profile.

The improvement is carried out by means of an iterative Least Mean Square (LMS) algorithm applied to each stroke. As a movement in a virtual target point modifies the entire trajectory, if we adapt all of them at the same time the iterative algorithm will not converge. For the sake of convergence, the refinement has to be carried out stroke by stroke in the same order than the original movement.

The refinement of the $j^{th}$ virtual target point is done as follows.
1. Set $j = 1$.
2. Refinement of the $j^{th}$ virtual target point
3. Work out the error or difference $\vec{dv_j}$ between the original and reconstructed trajectory salient points $\vec{dv_j} = sp_j - spr_j$.
4. Update the virtual target point as $tp_j = tp_j + \mu \vec{dv_j}$, being $0 < \mu < 1$.
5. With this new virtual target point, reconstruct the entire trajectory $T_r(t) = \{x_r(t), y_r(t)\}$ and work out the new salient points $spr_j, j = 1, ..., N-1$.
6. Set $j = j + 1$ and go to 2 if $j < N$.

An illustration of the result of this procedure can be seen in Figure 5. As it can be seen, the improvement of the reconstructed trajectory and velocity is significant. The improvement of the refinement over a signature is shown in Figure 4.

This refinement can be repeated as many time as necessary. In our experiments, the refinement hardly improves the SNR in trajectory and velocity after two full iterations with $\mu = 1$.

*G. Discussion*

It is expected that the Sigma-Lognormal parameters describes the Neuromotor system. As during a particular rapid movement the muscular system barely changes, the values of the parameters should somehow be stable or show a small variability. In this sense, Figure 6 shows the stroke time delay $\mu_j$ and the stroke time response $\sigma_j$ evolution. It should be highlighted the low variability of both paramters along the complex movement.

Additionally, the diferences between the start time of the stroke occurrence are values between 0.05 and 0.12 seconds which is coherent with the existence of the so called Central Pattern Generator (CPG). As it is suggested in [83], CPG produces rhythmic patterned outputs without sensory feedback. Moreover, it has been proposed that the mammalian locomotor CPG comprises a "timer" which generates step cycles of varying duration and a pattern formation layer which selects and grades the activation of motor pools. The step cycle is thought to be between 0.08 and 0.12 which roughly coincides with iDeLog $t_{0_j} - t_{0_{j-1}}$ values.

Finally, the value of the amplitude of the stroke $D_j$ follows the velocity as greater velocity in similar time results in wider strokes. Therefore, this result seems to follow the underlying biological motor system.

## IV. iDeLog ALGORITHM EXTENSIONS

The Sigma-Lognormal model describes analytically the handwritten trajectory as an overlap sum of circumference arcs and the velocity as a weighed sum of lognormals. Nevertheless, other theories have described the trajectories with other curves as splines and the velocity bells with other functions as Gaussian or Gamma function, among others, as it was mentioned in Section I. This section aims to bridge the gap between these theories and Sigma-Lognormal by including the possibility of selecting another curves between virtual target points. Such further curves leads to iDeLog to approach the velocity bell in the framework of the kinematic theory of rapid movements.

*A. Generalizing the trajectory between virtual target points*

The arc of circumferences that link virtual point exhibit several limitations in the reconstruction of complex movements. Specifically, there are several trajectories between velocity minima that cannot be explained by linking virtual target points with arc of circumferences. For instance, some trajectories between virtual target points includes inflection points, supposing a virtual target point per velocity minima. It can be generated by overlapping a convex and concave arcs of circumference. However, it is not the case of all the inflection points as can be seen in Figure 6.

In these cases, we have two possibilities for improving the reconstructed trajectory:

1) Adding a new target point around the inflection point linked with the neighbor ones with a concave and convex arcs of circumference. But, from the velocity point of view, this would require two extremely overlapped lognormal to fit the unique velocity bell. As a result, the concave and convex arc of circumference would be nearly full overlapped producing a straight line as trajectory. Therefore, this option is unsuitable.

2) Looking for a new ballistic trajectory that transverses the virtual target points $tp_{j-1}$ and $tp_j$ with tangent angle equal to $\theta_{s_j}$ and $\theta_{e_j}$ respectively. In this case, we should redefine Eq.(2) to (4) to reconstruct the trajectory as follows.

Let us define the new trajectory between virtual target points by a general parametric curve:

$$\begin{aligned} x &= f_j(u) \\ y &= g_j(u) \end{aligned} \quad (20)$$

which holds that $tp_{j-1} = \{f_j(u_{j1}), g_j(u_{j1})\}$, $tp_j = \{f_j(u_{j2}), g_j(u_{j2})\}$, its derivate in $u = u_{j1}$ is $\tan(\theta_{s_j})$ and its derivate in $u = u_{j2}$ is $\tan(\theta_{e_j})$. The length of the curve between the virtual targets points is:

$$D_j = \int_{u_{1j}}^{u_{2j}} \sqrt{\left(\frac{\partial f_j(u)}{\partial u}\right)^2 + \left(\frac{\partial g_j(u)}{\partial u}\right)^2}\, du \quad (21)$$

Accordingly to eq. (4), the trajectory can be reconstructed as:

$$\vec{s}(t) = \sum_{j=1}^{N} \begin{bmatrix} f_j(u_j(t)) \\ g_j(u_j(t)) \end{bmatrix} \quad (22)$$

where $u_j(t)$ is the value that solves the equation:

$$\int_{u_{1j}}^{u_j(t)} \sqrt{\left(\frac{\partial f_j(u)}{\partial u}\right)^2 + \left(\frac{\partial g_j(u)}{\partial u}\right)^2} \, du =$$
$$= \frac{D_j}{2}\left(1 + \text{erf}\left(\frac{\ln(t - t_{0_j}) - \mu_j}{\sigma_j \sqrt{2}}\right)\right) \quad (23)$$

The solution to this equation holds $u_{j1} \leq u_j(t) \leq u_{j2}$. In this case, instead of using eqs. (3) and (4) to reconstruct a trajectory, iDeLog would use eqs. (21) - (23). Figure 7 illustrate the underlying idea of this procedure.

### B. Trajectory reconstruction: from arc of circumferences to Clothoids

In order to generalize the curve that links the trajectory between virtual target points in the Sigma-Lognormal model, clothoids represents a good option owing to its biological meaning and the parameters by which are defined: their start and end points plus their start and end angles.

A Clothoid is a curve whose curvature changes linearly with its curve length (the curvature of a circular curve is equal to the reciprocal of the radius). In this way the transition from $\theta_{s_j}$ and $\theta_{e_j}$ is smoother than in the case of a circumference and can comprise a inflection point.

Moreover, This curve optimize the acceleration and jerk of the ballistic trajectory which is a characteristic of human being trajectories [84]. In fact, it has already been used for modelling Graffiti [73]. Clothoids are also commonly referred to as Spiros, Euler spirals, or Cornu spirals.

Clothoids are defined by the following system of ordinary differential equations [85]:

$$x = \int_0^u \cos(\pi v^2/2) dv$$
$$y = \int_0^u \sin(\pi v^2/2) dv \quad (24)$$

The process of obtaining solution $(x, y)$ of a Clothoid, given the two virtual target points $tp_{j-1}$ and $tp_j$ and the two angles $\theta_{s_j}$ and $\theta_{e_j}$, can be conducted using a package software such as [86].

Thus, iDeLog would reconstruct the trajectory by using Clothoids. Formally, after extracting the number of strokes and estimating the velocity parameters $t_{0_j}$, $\mu_j$ and $\sigma_j^2$, the virtual target points $tp_j$ are worked out.

Instead of working out the angles through the circumference that transverse the points $sp_{j-1}$, $mp_j$ and $sp_j$, the following procedure was carried out to estimate the initial start $\theta_{s_j}$ and end $\theta_{e_j}$ angles:

1. The angle $\theta_{s_j}$ is obtained as the angle at $sp_{j-1}$ of the circumference that transverses the points $sp_{j-1}$, $mp_{j1}$ and $mp_j$, $mp_{j1}$ being the point in the middle of the trajectory between $sp_{j-1}$ and $mp_j$.
2. The angle $\theta_{e_j}$ is obtained as the angle at $sp_j$ of the circumference that transverses the points $mp_j$, $mp_{j2}$ and $sp_j$, $mp_{j2}$ being the point in the middle of the trajectory between $mp_j$ and $sp_j$.

The next step is to work out $D_j$ and the Clothoids that link the virtual target points with eq. (21), (22), (23) and (24). The refinement algorithm is performed in the same way. As this procedure is applicable to whatever curve in the form of eq. (20), obviously, it can also be used with arc of circumferences.

As an example, Figure 8 illustrates poorer reconstruction rates in trajectory when arcs of circumferences are used in complex strokes against the improvement with Clothoids.

### C. Generalization of the velocity bell shape

In motor control theory, various computational models have been developed to describe the velocity profiles of rapid movements [87]. Among them, the Kinematic Theory and the Minimization Theory have put forward analytical expressions to describe the velocity profiles of rapid movements. Both theories share a common framework in which motoneuron commands go through a set of $L$ subsystems with different impulse response. Thus, the bell-shaped pattern of a velocity profile emerges from the synergetic coupling of numerous neuromuscular subsystems [88].

Given the amount of different proposed shapes for the velocity profile [87] and the similarity among them, the next bell functions have been implemented in iDeLog:

#### 1) Gaussian function

Whether all the subsystems are considered independent, based on the central limit theorem, the bell-shaped velocity profile can be approached by a symmetric Gaussian when $L \to \infty$. The Gaussian function to model the velocity bell is defined as:

$$v_j(t; \mu_j, \sigma_j^2) = \frac{D_j}{\sigma_j \sqrt{2\pi}} \exp\left\{-\frac{(t - \mu_j)^2}{2\sigma_j^2}\right\} \quad (25)$$

where $D_j$ is the area of the velocity bell, $\mu_j$ is the mean and $\sigma_j^2$ is the variance. It is expected that the mean of the Gaussian function will be around the peak of the velocity bell. This function is supported in $[-\infty, +\infty]$ and it is symmetric.

#### 2) Gamma function

It is well-known that the velocity bells are asymmetric and they start at time $t_{0_j}$. As such, the minimum time theory model represents the architecture of the neuromuscular system by the convolution product of first order low-pass filters. When the number of low-pass filters tend to a large number, the velocity profile tends to a special case of Gamma function [88] which is asymmetric. Specifically, the Gamma function used to fit the velocity bell is defined by:

$$v_j(t; t_{0_j}, \alpha, \beta) = D_j \frac{(t - t_{0_j})^{\alpha-1} \exp\{-(t - t_{0_j})/\beta\}}{\beta^\alpha \int_0^\infty t^{\alpha-1} \exp\{-t\} dt} \quad (26)$$

where $\alpha$ and $\beta$ are called the shape and scale parameter respectively. This function is defined for $t_{0_j} < t < \infty$ and $\alpha, \beta > 0$. From the computational point of view, it adds a new parameter or freedom degree for the velocity bell fitting.

#### 3) Beta function

The minimum jerk theory generalizes the expression for the velocity bell as a smooth Beta model which is itself a particular case of Gamma function [88]. In this case the velocity bell is double bounded, i.e. it start at $t_{0_j}$ and ends at $t_{e_j}$. The beta function is defined by:

$$v_j(t; t_{0_j}, \alpha, \beta) = D_j \frac{(t-t_{0_j})^{\alpha-1}(1-(t-t_{0_j}))^{\beta-1}}{\int_0^1 u^{\alpha-1}(1-u)^{\beta-1}du} \quad (27)$$

where $\alpha$ and $\beta$ are the first and second shape parameter. This function is defined for $t_{0_j} < t < t_{e_j} = t_{e_j} + 1$, and $\alpha, \beta > 0$. Due to the inherent finite time length of the beta velocity bell, it is required to be hold that $t_{min_j} - t_{0_j} < 1$, which is not a practical problem if $t_{min,j-1} - t_{0_j} \approx 0.5$ as suggested in Section III.B.1.

*4) Lognormal function*

However, neuromuscular subsystems are physically connected to each other and the system dependency cannot be neglected. In a further step, the Kinematic Theory takes such dependency into account. This way, it uses the Central Limit Theorem, which shows that the velocity shape tends to a lognormal impulse response when the number of subsystems tends toward infinite.

The lognormal function has been defined in eq. (1) and its span is $t_{0_j} < t < \infty$.

*5) Double-Bounded lognormal*

This function introduces an extension to the infinite length response lognormal to allow for both a lower and an upper bound to the values of the lognormal. This extension is called the four parameter distribution in [89] or the double bounded lognormal in [87] and confines the lognormal to the range $t_{0_j} < t < t_{e_j}$ by assuming that $(t-t_{0_j})/(t_{e_j}-t)$ is lognormal. As a result, the double bounded lognormal is defined as:

$$v_j(t; t_{0_j}, t_{e_j}, \mu_j, \sigma_j^2) = \frac{D_j(t_{e_j}-t_{0_j})}{\sigma_j \eta(t)\sqrt{2\pi}} \exp\left\{-\frac{[\ln \zeta(t) - \mu_j]^2}{2\sigma_j^2}\right\} (28)$$

where $\zeta(t) = (t-t_{0_j})/(t_{e_j}-t)$ and $\eta(t) = (t-t_{0_j})(t_{e_j}-t)$. The variables $D_j$, $t_{0_j}$, $\mu_j$, and $\sigma_j^2$ are named after the lognormal and $t_{e_j}$ is the end time of the lognormal. In this case, we have to estimate an additional parameter, i.e. $t_{e_j}$. It is, therefore, an additional freedom degree that leads to a better adjustment of the velocity bell.

*6) Generalized Extreme Value function (GEV)*

The different velocity profile models are defined when the number of subsystem (i.e. number of muscles involved in the human action) is large, specifically when tends to infinite. As a result, it could be said that the velocity bell tends to lognormal but there could be some deviations depending on the no infinite number of muscles involved in the movement and their correlation.

In this sense, iDeLog provide the possibility of fitting the velocity bell with a Generalized Extreme Value (GEV) function. GEV has been successfully applied to model physical and biological phenomena. It appears as a particular case of a more general problem of limit theorem for sums of strongly correlated systems [90]. GEV combines three simple distributions into a single form, allowing a continuous range of possible shapes. As a consequence, the GEV leads to "let the data decide" which distribution is most appropriate. The GEV is defined as:

$$v_j(t; t_{0_j}, \xi_j, \mu_j, \sigma_j^2) = \frac{D_j}{\sigma_j} s(t-t_{0_j})^{\xi_j+1} e^{-s(t-t_{0_j})} \quad (29)$$

where

$$s(t) = \begin{cases} \left[1 + \xi_j \left(\frac{t-\mu_j}{\sigma_j}\right)\right]^{-1/\xi_j} & if\ \xi_j \neq 0 \\ exp\{-(t-\mu_j)/\sigma_j\} & if\ \xi_j = 0 \end{cases} \quad (30)$$

$\mu_j$ being the location parameter, $\sigma_j$ the scale parameter and $\xi_j$ the shape parameter. The shape parameter governs the tail behavior and identifies the bell as belonging to one of the three sub-families of distributions:

1. If $\xi_j = 0$ is a Gumbel distribution which is unbounded.
2. If $\xi_j > 0$, is a Fréchet distribution, with lower tail bounded and heavy upper tail.
3. If $\xi_j < 0$, is a Weibull distribution, with upper tail bounded and short.

As can be seen, it has the same number of parameters than the double bounded lognormal.

An example of velocity profiles reconstructed with each of the velocity bell models is shown in Figure 9.

V. EXPERIMENTAL ASPECTS ON RAPID MOVEMENTS

The evaluation aims to show the ability of iDeLog to obtain a valid set of Sigma-Lognormal parameters. In this way, two experiments have been carried out:

- **Experiment 1: Script-Studio and iDeLog.** A comprehensive comparison between Script-Studio and iDeLog is discussed. It is a first requirement to analyze the convenience of using iDeLog for modeling human rapid movements.
- **Experiment 2: Analysis of iDeLog novelties.** To provide the results of iDeLog with the new extensions of the algorithm, i.e., using Clothoids instead of circumference arcs or the double-bounded lognormal instead of the lognormal and so on. In this experiment, we provide quantitative measures to each novel functionality in iDeLog.

These experiments have been conducted with three different databases and evaluated with well-known measures.

*A. Databases*

As iDeLog is a novel implementation of Theory of rapid movements, we have used handwriting signatures. They are supposed to be executed rapidly involving a large number of muscles. For these purposes, the experiments are carried out with the genuine signatures of three publicly available on-line signature databases. We have selected the genuine signatures as all of them are well-learned movements written naturally. In this way we avoid biased measures owing to constrained and sluggish movements of imitated signatures. Regarding to the genuine signatures, we describe the databases used as follows:

- BiosecureID-Signature UAM subcorpus [91]. It is a subset of a multimodal database. It consists of 132 users. Each user provided for 4 genuine signatures in each session. In total, were recorded 4 sessions distributed in a 4-month time span. The signatures were acquired by a WACOM tablet.
- MCYT100 sub-corpus [92]. They are the 100 first

signatures of the full MCYT-330 database which it was captured by a WACOM tablet. It contains 25 genuine signatures per user acquired in two sessions.

- SUSIG-Visual sub-corpus [93]. This database contains 94 users with 20 genuine signatures per user, acquired in two sessions. This sub-corpus was collected with an LCD touch device. As result, there are some jitter in the sampling period which leads to noisy lognormals.

### B. Quantitative Measures

The reconstruction results are given in terms of Signal to Noise Ratio ($SNR$), Segmented Signal to Noise Ratio ($SNRSeg$) and the ratio of the $SNR$ and the number of lognormal detected ($SNR/NbLog$) for trajectory and velocity.

The Signal to Noise Ratio for trajectory $SNR_t$ measures the difference between the original and reconstructed trajectory. In case of preprocessing the original trajectory, the measure compares the original preprocessed trajectory with the reconstructed one. It is defined as:

$$SNR_t = 10\log\left(\frac{\int_{t=0}^{T}(x_{op}(t)^2 + y_{op}(t)^2)dt}{\int_{t=0}^{T}(dx_{or}(t)^2 + dy_{or}(t)^2)dt}\right) \quad (31)$$

being in the numerator $x_{op}(t) = x_o(t) - \overline{x_o(t)}$, $y_{op}(t) = y_o(t) - \overline{y_r(t)}$ and in the denominator $dx_{or}(t) = x_o(t) - x_r(t)$, $dy_{or}(t) = y_o(t) - y_r(t)$. The subindexes $o$ and $r$ denote respectively the original and the reconstructed signal.

The Signal to Noise Ratio for velocity $SNR_v$ measure is the one used by the Script-Studio. This measure is considered to index the lognormality of the movements. As these experiments deal with genuine signatures which are supposed to be lognormal as they are handwritten fluently, the higher the $SNR$ obtained, the better the reconstruction. The SNR is defined as:

$$SNR_v = 10\log\left(\frac{\int_{t=0}^{T} v_o(t)^2 \, dt}{\int_{t=0}^{T}(v_o(t) - v_r(t))^2 dt}\right) \quad (32)$$

The segmented Signal to Noise Ratio for trajectory $SNRseg_t$ is an alternative measure to the $SNR_t$ with a similar meaning. It is defined as the average of the $SNR_t$ obtained in every stroke along the signature, assuming as a stroke the trace between two consecutive velocity minima. This measure avoids the bias of the $SNR_t$ when the signature comprises short and long strokes. In this case, the bias is due to high or low $SNR_t$ in long strokes overrule bad or good fitting in the remainder short strokes. The $SNRseg_t$ is calculated as:

$$SNRseg_t = \frac{10}{N}\sum_{j=1}^{N-1}\log\left(\frac{\int_{t_{min_{j-1}}}^{t_{min_j}}(x_{op}(t)^2 + y_{op}(t)^2)dt}{\int_{t_{min_{j-1}}}^{t_{min_j}}(dx_{or}(t)^2 + dy_{or}(t)^2)dt}\right) \quad (33)$$

For the case of the velocity, the segmented Signal to Noise Ratios $SNRseg_v$ is defined as

$$SNRseg_v = \frac{10}{N}\sum_{j=1}^{N-1}\log\left(\frac{\int_{t_{min_{j-1}}}^{t_{min_j}} v_o(t)^2}{\int_{t_{min_{j-1}}}^{t_{min_j}}(v_0(t) - v_r(t))^2}\right) \quad (34)$$

Finally, the ratio of the $SNR_v$ and the number of lognormal detected $SNR_v/NbLog$ has been proposed as a variable that reflects the writer's ability to make regular movements. It is a good global indicator of the graphomotor performance of a writer [94][95]. Instead, in comparing two algorithms, the variable index the number of lognormal required by the algorithm to reach the given $SNR_v$. Moreover, the more $SNR_v/NbLog$, the better the algorithm performance. Similarly is defined the $SNR_t/NbLog$

The quantitative measures described in this section were obtained for each single specimens. The averaged result is provided for each experiment.

### C. Experiment 1: Script-Studio and iDeLog comparison

For a fair comparison between Script-Studio and iDeLog, the original signal used as input to both algorithms should be the same. Consequently, the input to iDeLog is interpolated and smoothed, the virtual target points are linked with arc of circumferences and the velocity bells are approached with lognormals probability distributions. The results are shown in Table 1. As can be seen, the performance of iDeLog fitting the velocity profile is slightly lower than Scrip-Studio but the adjustment of the trajectory is significantly better in all the cases. The slight decrease in $SNR_v$ of iDeLog is due to the constraints fitting he trajectory.

*Table 1. Results of Scrip-Studio and iDeLog with and without preprocessing, linking virtual target points with arcs of circumference and modelling velocity bells with lognormal.*

| Database | Database | Preprocessing | $SNR_t$ | $SNRseg_t$ | $SNR_t/NbLog$ | $SNR_v$ | $SNRseg_v$ | $SNR_v/NbLog$ |
|---|---|---|---|---|---|---|---|---|
| BiosecureID | ScriptStudio | YES | 6.47dB | 0.36dB | 0.283 | 16.35dB | 16.18dB | 0.740 |
| | iDeLog | YES | 23.28dB | 22.71dB | 0.880 | 15.22dB | 15.32dB | 0.569 |
| | iDeLog | NO | 22.63dB | 19.36dB | 0.881 | 16.20dB | 16.04dB | 0.579 |
| MCYT100 | ScriptStudio | YES | 6.57dB | 2.24dB | 0.298 | 16.02dB | 16.57dB | 0.698 |
| | iDeLog | YES | 21.33dB | 22.74dB | 0.905 | 15.22dB | 15.46dB | 0.630 |
| | iDeLog | NO | 20.81dB | 15.66dB | 0.748 | 15.31dB | 15.26dB | 0.537 |
| SUSIG Visual | ScriptStudio | YES | 16.39dB | 12.38dB | 0.808 | 17.78dB | 19.37dB | 0.881 |
| | iDeLog | YES | 24.11dB | 24.02dB | 1.338 | 15.19dB | 15.34dB | 0.841 |
| | iDeLog | NO | 21.63dB | 10.82dB | 0.709 | 10.25dB | 12.50dB | 0.358 |

### D. Experiment 2: iDeLog extension

This section shows some results using the available extensions in iDeLog such as switching off the preprocessing in the original signals, linking the virtual target points with Clothoids instead of with arcs of circumference and approaching the velocity bell with different probability distributions.

#### 1) Without preprocessing the original signal.

The iDeLog algorithm allows to decompose the original signal without the default preprocessing of Script-Studio. The results obtained are summarized in the no preprocessing cases in Table 1. Obviously, the results with the preprocessed databases are better since the signal to decompose is considerably less noisy. Notice that in the preprocessing case, the SNRs are obtained comparing the reconstructed signal with the original one preprocessed which is not the real one.

Furthermore, the results with BiosecureID and MCYT100 databases are not too different with and without preprocessing, but there are significant differences with SUSIG Visual database. The main reason is the noisy velocity profiles of SUSIG Visual possibly due to irregularities in the sampling frequency as it is acquired with a touch tablet instead of a

*Table 3. Results of iDeLog using clothoids as link between virtual target points and Lognormal as velocity bell function*

| Database | Smoothing | Link between target points | $SNR_t$ | $SNRseg_t$ | $SNR_t/NbLog$ | $SNR_v$ | $SNRseg_v$ | $SNR_v/NbLog$ |
|---|---|---|---|---|---|---|---|---|
| BiosecureID | YES | Circular | 23.28dB | 22.71dB | 0.880 | 15.22dB | 15.32dB | 0.569 |
| | | Clothoid | 26.62dB | 25.15dB | 1.005 | 15.47dB | 15.87dB | 0.579 |
| | NO | Circular | 22.63dB | 19.36dB | 0.881 | 16.20dB | 16.04dB | 0.579 |
| | | Clothoid | 26.00dB | 21.64dB | 0.939 | 16.69dB | 16.74dB | 0.597 |
| MCYT100 | YES | Circular | 21.33dB | 22.74dB | 0.905 | 15.22dB | 15.46dB | 0.630 |
| | | Clothoid | 24.65dB | 25.02dB | 1.032 | 15.48dB | 16.00dB | 0.642 |
| | NO | Circular | 20.81dB | 15.66dB | 0.748 | 15.31dB | 15.26dB | 0.537 |
| | | Clothoid | 22.37dB | 18.46dB | 0.888 | 15.49dB | 15.28dB | 0.594 |
| SUSIGVisual | YES | Circular | 24.11dB | 24.02dB | 1.338 | 15.19dB | 15.34dB | 0.841 |
| | | Clothoid | 28.43dB | 26.84dB | 1.575 | 15.55dB | 16.03dB | 0.860 |
| | NO | Circular | 21.63dB | 10.82dB | 0.709 | 10.25dB | 12.50dB | 0.358 |
| | | Clothoid | 23.80dB | 14.31dB | 0.788 | 10.04dB | 12.99dB | 0.355 |

WACOM. This fact is detected not only by the low $SNR_v$ but also by the reduced $SNRseg_t$.

*2) Using Clothoids instead of circumference arcs to link virtual target points in the trajectory*

The improvement of using Clothoids instead of arcs of circumference is shown in Table 2. As it can be seen in all cases, excellent improvements in reconstructing the trajectory are obtained as well as a slight improvement in the velocity reconstruction, except on the case of noisy velocity profiles (SUSIGVisual dataset without preprocessing).

*3) Different velocity bell functions*

The iDeLog algorithm includes the possibility of reconstructing the rapid movement with different velocity bell. The results of this possibility are shown in Table 3. As a rule of thumb, our findings are:
1. Using different velocity bell function, the reconstruction of the trajectory is nearly the same.
2. The performance of the velocity fitting depends on the velocity bell function. The Gaussian, Lognormal and Gamma functions provide similar results. The Beta function slightly improves the fitting. But the Double Bounded Lognornal and GEV function, with four parameters to fit the velocity bell, significantly and consistently outperform the previous results in all the cases.

## VI. CONCLUSIONS

This paper presents a new procedure, called iDeLog, to work out the parameters of the Sigma-Lognormal model. As novelty, iDeLog reconstructs trajectory and velocity of a movement at the same time. Moreover, iDeLog includes the following characteristics: 1) possibility of avoiding the initial preprocessing, 2) possibility of linking the virtual target points with circumference arcs and Clothoids and 3) possibility of model the velocity bell with the next functions: Gaussian, Lognormal, Gamma, Beta, Double Bounded Lognornal and GEV.

*Table 2. iDeLog result without preprocessing the original signal, using Clothoids as link between virtual target points and different velocity bells.*

| Database | Velocity Bell | $SNR_t$ | $SNRseg_t$ | $SNR_t/NbLog$ | $SNR_v$ | $SNRseg_v$ | $SNR_v/NbLog$ |
|---|---|---|---|---|---|---|---|
| BiosecureID | Gaussian | 26.17dB | 21.54dB | 0.945 | 17.08dB | 16.96dB | 0.611 |
| | Lognormal | 26.00dB | 21.64dB | 0.939 | 16.69dB | 16.74dB | 0.597 |
| | Gamma | 26.07dB | 21.46dB | 0.941 | 16.90dB | 16.76dB | 0.605 |
| | Beta | 26.13dB | 21.51dB | 0.944 | 17.21dB | 17.00dB | 0.616 |
| | DBLognormal | 26.01dB | 21.59dB | 0.941 | 18.77dB | 18.01dB | 0.673 |
| | GEV | 25.88dB | 20.97dB | 0.934 | 19.36dB | 18.58dB | 0.695 |
| MCYT100 | Gaussian | 22.48dB | 18.51dB | 0.894 | 15.89dB | 15.58dB | 0.611 |
| | Lognormal | 22.38dB | 18.46dB | 0.888 | 15.49dB | 15.28dB | 0.594 |
| | Gamma | 22.39dB | 18.47dB | 0.891 | 15.67dB | 15.41dB | 0.601 |
| | Beta | 22.48dB | 18.49dB | 0.893 | 15.93dB | 15.58dB | 0.612 |
| | DBLognormal | 22.33dB | 18.76dB | 0.888 | 17.01dB | 16.39dB | 0.649 |
| | GEV | 21.95dB | 17.85dB | 0.876 | 16.61dB | 16.11dB | 0.638 |
| SUSIGVisual | Gaussian | 23.94dB | 14.27dB | 0.796 | 10.21dB | 13.17dB | 0.361 |
| | Lognormal | 23.80dB | 14.31dB | 0.788 | 10.04dB | 12.99dB | 0.355 |
| | Gamma | 23.69dB | 14.28dB | 0.785 | 10.11dB | 13.07dB | 0.358 |
| | Beta | 23.86dB | 12.22dB | 0.792 | 10.20dB | 13.18dB | 0.360 |
| | DBLognormal | 25.09dB | 14.56dB | 0.820 | 15.76dB | 14.10dB | 0.525 |
| | GEV | 23.39dB | 13.76dB | 0.773 | 9.867dB | 13.49dB | 0.353 |

DBLognormal: Double Bounded Lognormal function

*Table 4. Glossary of variables*

| Symbol | Meaning |
|---|---|
| $N$ | Number of virtual target points |
| $T$ | Temporal length of the signature |
| $\{D_j, t_{0_j}, \mu_j, \sigma_j^2\}_{j=1}^N$ | Sigma Lognormal parameters of stroke $j$ |
| $\{\theta_{s_j}, \theta_{e_j}\}_{j=1}^N$ | Start and end angle of the link between $tp_{j-1}$ and $tp_j$ |
| $\{tp_j\}_{j=0}^N$ | Virtual Target point of stroke $j$ |
| $\{sp_j\}_{j=0}^N$ | Salient points (velocity minima) of original signature |
| $\{spr_j\}_{j=0}^N$ | Salient points of reconstructed trajectory. |
| $\{t_{min_j}\}_{j=0}^N$ | Time of the salient points $\{sp_j\}_{j=0}^N$ and $\{spr_j\}_{j=0}^N$ |
| $\{x_o(t), y_o(t)\}_{t=0}^T$ | Samples of the online original trajectory |
| $\{x_r(t), y_r(t)\}_{t=0}^T$ | Samples of the online reconstructed trajectory |
| $v_o(t)$ | Velocity profile of the original movement |
| $v_r(t)$ | Velocity profile of the reconstructed movement |
| $\{v_j(t)\}_{j=1}^N$ | Velocity profile of stroke $j$, $0 < t < T$ |
| $SNR_v, SNR_t$ | Signal to Noise Ratio between original and reconstructed velocity and trajectory respectively |
| $SNRseg_v, SNRseg_t$ | Segmented Signal to Noise Ratio between original and reconstructed velocity and trajectory. |

The experiments conducted show that ScripStudio displays a slight better velocity adjustment than iDeLog in the case of linking virtual target point with circumference arc and modelling the velocity bell with lognormal functions but iDeLog outperforms significantly in the case of trajectory. The barely lower velocity adjustment is attributed to the tradeoff of adjusting trajectory and velocity at the same time.

Script-Studio uses arcs of circumferences and lognormals probability distributions for reconstructing trajectories and velocity profiles respectively. Nevertheless, using Clothoids and four parameters velocity bell functions in iDeLog, we can improve meaningfully the performance in terms of $SNR$.

The algorithm iDeLog is freely distributed as a Matlab toolbox.

The new procedure is expected to be a further step for applications based on the Sigma-Lognornal model and the Kinematic Theory of rapid movements. Moreover, it can be specially promising in those cases in which there are some information in high frequency components which are removed in the preprocessing stage. On the other hand, the comparison after reconstructing the same original signal with and without preprocessing, with circumference arcs or Clothoids and different velocity bell functions will provide new insights into the human movement analysis.

New version of iDeLog are being undertaken for working out the Sigma-Lognormal parameters of 3D trajectories and 8-connected trajectories without temporal information.

ACKNOWLEDGMENT

This study was funded by the Spanish government's MIMECO TEC2012-38630-C04-02 research project and European Union FEDER program/funds.

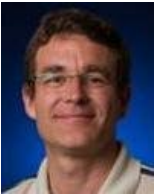

**Miguel A. Ferrer** received the M.Sc. and Ph.D. degrees from the Universidad Politécnica de Madrid, Madrid, Spain, in 1988 and 1994, respectively. He joined the University of Las Palmas de Gran Canaria, Las Palmas, Spain, in 1989, where he is currently a Full Professor. He established the Digital Signal Processing Research Group in 1990. His current research interests include pattern recognition, biometrics, audio quality, and computer vision applications to fisheries and aquaculture.

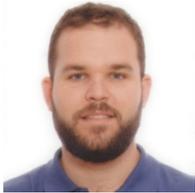

**Moises Diaz** (M'15) received the M.Tech., M.Sc., and Ph.D. degrees in engineering, and the M.Ed. degree in secondary education from La Universidad de Las Palmas de Gran Canaria, Las Palmas, Spain, in 2010, 2011, 2016, and 2013, respectively. He is currently an associate professor at Universidad del Atlantico Medio, Spain. His current research interests include pattern recognition, document analysis, handwriting recognition, biometrics, computer vision, and intelligent transportation systems.

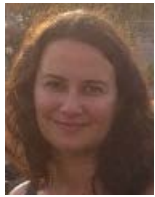

**Cristina Carmona-Duarte** received the Telecommunication Engineering degree in 2002 and de Ph.D. degree in 2012 from Universidad de Las Palmas de Gran Canaria. She has been assistant Professor at Universidad de Las Palmas where she is currently a researcher. Her research areas include high resolution radar, pattern recognition and biometrics.

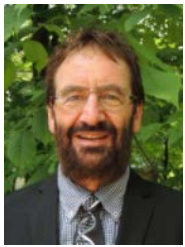

Réjean Plamondon received a B.Sc. degree in Physics, and M.Sc.A. and Ph.D. degrees in Electrical Engineering from Université Laval, Québec, P.Q., Canada in 1973, 1975 and 1978 respectively. In 1978, he joined the faculty of the École Polytechnique, Université de Montréal, Montréal, P.Q., Canada, where he is currently a Full Professor. He has been the Head of the Department of Electrical and Computer Engineering from 1996 to 1998 and the President of Ecole Polytechnique from 1998 to 2002. He is now the Head of Laboratoire Scribens at this institution.